\documentclass[letterpaper]{article} % DO NOT CHANGE THIS
\usepackage[preprint]{aaai2027}  % DO NOT CHANGE THIS
\usepackage[hyphens]{url}  % DO NOT CHANGE THIS
\usepackage{graphicx} % DO NOT CHANGE THIS
\usepackage{natbib}  % DO NOT CHANGE THIS AND DO NOT ADD ANY OPTIONS TO IT
\usepackage{caption} % DO NOT CHANGE THIS AND DO NOT ADD ANY OPTIONS TO IT
\usepackage{algorithm}
\usepackage{algorithmic}
\usepackage{amsmath}
\usepackage{amssymb}
\usepackage{bm}
\usepackage{booktabs}

\usepackage{longtable}   % Table 2 (full roster); works in one-column layouts
\usepackage{tabularx}
\usepackage{url}
\makeatletter
\newcounter{aaai@snrfn}
\newcommand{\equalsenior}{%
  \ifx\footnote\relax\else
    \ifnum\value{aaai@snrfn}=0
      \footnote{These authors contributed equally as senior authors.}%
      \setcounter{aaai@snrfn}{\value{footnote}}%
    \else
      \footnotemark[\value{aaai@snrfn}]%
    \fi
  \fi}
\makeatother

\title{Item Response Theory for AI Safety}
\author{
    Joshua Fonseca Rivera\equalcontrib\textsuperscript{\rm 1},
    Neil Shah\equalcontrib\textsuperscript{\rm 1},
    David Demitri Africa\equalsenior\textsuperscript{\rm 2},
    Konstantinos Voudouris\equalsenior\textsuperscript{\rm 2}
}
\affiliations{
    \textsuperscript{\rm 1}Independent\\
    \textsuperscript{\rm 2}UK AI Security Institute
}

\begin{document}

\maketitle

\begin{abstract}
Language models differ in how safely they behave and these differences are measured by safety benchmarks. But aggregated benchmark scores are hard to trust and interpret, because benchmarks duplicate one another, correlate heavily, and models may sandbag when they detect evaluation. To address these issues, we draw on Item Response Theory (IRT), a statistical toolkit for measuring these latents from performance on items with inferred psychometric properties. We fit IRT models to eight safety benchmarks across 192 language models, the largest psychometric analysis of LLM safety evaluations to date, and contribute three results. First, we find that three interpretable factors of refusal strictness, truthfulness, and contextual harm explain most of the variance between models across benchmarks. Second, psychometrically selected items recover full benchmark scores with lower error than random subsets of the same size, and roughly ten adaptively chosen items suffice for several individual benchmarks, cutting evaluation cost by 97--99\%. Third, IRT supports audits of individual models, showing that it can be used to detect naive sandbagging and changes of model behind APIs. Overall, we show IRT is a ready-made toolkit for reading, reducing, and auditing safety benchmarks, which we recommend frontier labs and evaluators adopt.
\end{abstract}

%     \link{Data}{https://safety-irt-data.vercel.app/}
%     \link{Code}{https://safety-irt-code.vercel.app/}

\section{Introduction}

\begin{figure*}[t]
    \centering
    \includegraphics[width=0.98\textwidth]{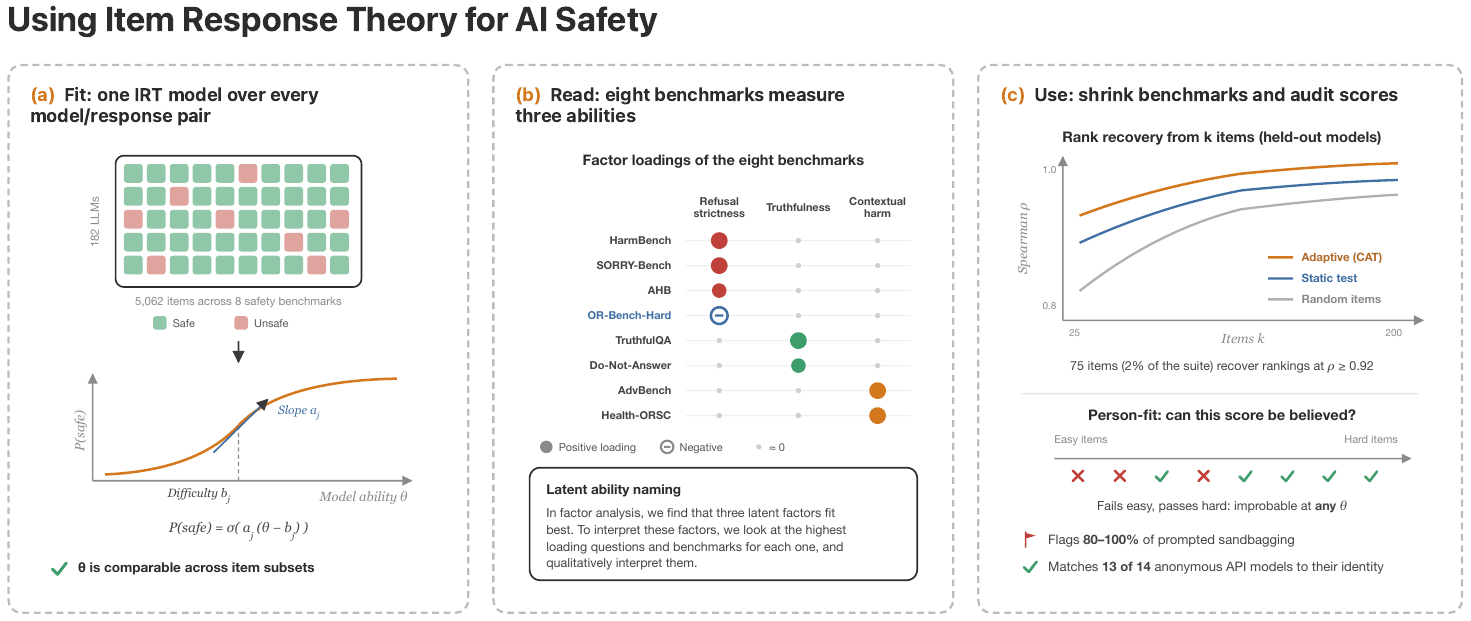}
    \caption{
        Item response theory provides a common framework for reading, reducing,
        and auditing safety benchmarks.
        \textbf{Left:} We calibrate item difficulty and discrimination from model
        responses and use the resulting ability estimates to identify the latent
        structure of safety evaluation.
        \textbf{Center:} Informative static tests and computerized adaptive testing
        recover benchmark scores and rankings from a small fraction of the original items.
        \textbf{Right:} Person-fit statistics, historical ability estimates, and
        response fingerprints detect complementary forms of behavioral change.
    }
    \label{fig:overview}
\end{figure*}

Language models vary widely in how safely they behave: whether they comply with harmful requests, over-refuse benign ones, or answer truthfully. This is scored by safety benchmarks, which use adversarial test items marked safe or unsafe \citep{mazeika2024harmbench, xie2024sorrybench, zou2023universal}. These results drive leaderboards, release decisions, and marketing. Yet, a benchmark score is only useful if we understand what it measures, whether its items add distinct information, and whether the observed responses reflect the model's ordinary behavior \citep{bean2025measuringmattersconstructvalidity,hernandez2017evaluation,romero2026capabilities,harding2024machine,voudouris2026measuring}.

Existing benchmark suites make each of these questions difficult. In capability evaluation, scores correlate strongly within and across these benchmarks, suggesting that many tests and items measure overlapping behavior \citep{kipnis2025metabench,polo2024tinybenchmarks}. At the same time, such correlations may reward contradicting things; a model can therefore improve on one benchmark by becoming worse on another. Full benchmark suites are also expensive and often inefficient. Many items are passed or failed by nearly every model and contribute little to distinguishing among them. Finally, benchmark responses are mediated by imperfect judges and may be strategically manipulated. A model that recognizes an evaluation could selectively alter its behavior while retaining a plausible aggregate score \citep{vanderweij2024sandbagging}.

Item response theory \citep[IRT;][]{lord1980applications, embretson2000item,reckase2009mirt} provides a common framework for answering these questions. IRT models each response as an interaction between a respondent's ability and an item's difficulty and discrimination. Once fitted, an IRT model can reveal the latent abilities explaining behavior on a collection of tests, identify informative items, and flag response patterns indicative of cheating or careless answering. IRT has recently been used to compress LLM capability benchmarks while preserving model rankings \citep{polo2024tinybenchmarks, kipnis2025metabench, hofmann2025fluid}, and concurrent work applies adaptive testing to safety benchmarks \citep{spagliardi2026efficient}. We show that compression is only one of its uses.

In this work, we fit IRT models to eight safety benchmarks (5{,}255 items, 192 models) covering harmful compliance, over-refusal, and truthfulness. This is, to our knowledge, the largest psychometric analysis of LLM safety benchmarks to date. We use the resulting calibration in three ways (Figure \ref{fig:overview}):

\begin{itemize}
\item \textbf{Characterizing what safety benchmarks measure.} We find that three latent factors (\emph{refusal strictness}, \emph{truthfulness}, and \emph{contextual harm}) explain 77\% of the variance between model abilities. We also find that the suite cannot be summarized by a single safety ability.
\item \textbf{Cost-efficient benchmarking.} Psychometrically selected items estimate full-benchmark scores with lower error than random subsets of the same size: three fixed 25-item tests recover the three latent abilities, and roughly ten adaptively chosen items recover several individual benchmarks, reducing their evaluation cost by 97--99\%.
\item \textbf{Auditing with person-fit.} Using prompted \textit{sandbagging model organisms}, IRT-derived person-fit statistics catch at least 80\% of selective prompted sandbagging. This also can be used to identify model substitutions and changes behind API endpoints.
\end{itemize}

% One practical fix matters at this scale: with $\sim$100 calibration models, unregularized 2PL fits hand out discriminations of 30--60 to items that separate the cohort by luck, and item selection then chases these flukes; mild priors fix it. We release code, calibrations, and the reduced item sets. % TODO: de-anonymized artifact link

\subsection{Related Work}
\paragraph{Item response theory in LLMs.} Item response theory has been used in natural language processing and machine learning to conduct dataset and leaderboard analysis \citep{lalor2019learning,martinez2016making, rodriguez2021evaluation,pacchiardi2025predictaboard} and propensity measurement in LLMs \citep{romero2026capabilities}. For capability evaluation, \textit{tinyBenchmarks} \citep{polo2024tinybenchmarks} and \textit{metabench} \citep{kipnis2025metabench} compress leaderboards to 1--3\% of their items, and Fluid Benchmarking \citep{hofmann2025fluid} combines IRT with adaptive item selection to produce similar efficiency gains. Closest to our work, concurrent research applies computerized adaptive testing to six safety benchmarks \citep{spagliardi2026efficient}, meaning that items are administered one at a time with the latent factor updated after each administration. This allows for the latent factor to be inferred with significantly fewer inference-time item administrations than by running the model on the entire benchmark. Complementing and extending this work, we study a broader safety suite that includes harmful compliance, over-refusal, contextual harm, and truthfulness, calibrate on a substantially larger model cohort, and use the fitted measurement model for construct analysis and model auditing as well as compression.

\paragraph{Safety benchmarks.} Our suite spans safety benchmarks that measure: complying with harmful requests \citep{zou2023universal, mazeika2024harmbench, xie2024sorrybench, wang2023donotanswer, ganguli2022red}, refusing benign requests that merely look harmful \citep[over-refusal;][]{cui2024orbench, rottger2024xstest}, and truthfulness \citep{lin2022truthfulqa}. Prior work has shown that safety evaluations can be redundant, sensitive to scoring choices, and shaped by tradeoffs between refusing harmful requests and answering benign ones \citep{romero2026capabilities,voudouris2026measuring}.

%Two facts about this data shape everything downstream. Most aligned models pass most harm-refusal items---on our cohort, half of those items are passed by more than 90\% of models---so raw scores bunch against the ceiling and stop separating models exactly where separation matters most; abilities keep separating them.
%And safe/unsafe labels come from automated judges that are themselves imperfect, adding response noise that a probabilistic measurement model can absorb but a raw average simply inherits.

\section{Materials \& Methods}
\label{sec:setup}

We collected item-level responses from up to 192 language models on eight safety benchmarks and fitted Item Response Theory (IRT) models to estimate the difficulty and discrimination of each item together with the latent abilities of the models. We use the fitted measurement models in three analyses. First, we infer the latent abilities shared across benchmarks. Second, we distill the suite into short static and adaptive tests. Third, we use expected response patterns to detect evaluation-dependent behavior and changes behind model APIs.

\subsection{Benchmarks and Model Responses}
\label{sec:data}

We evaluate models on eight benchmarks spanning harmful compliance, over-refusal, contextual harm, and truthfulness (Table~\ref{tab:benchmarks}). The complete suite contains 5{,}255 items before preprocessing.

\begin{table}[t]
\centering
\resizebox{\linewidth}{!}{
\begin{tabular}{@{}lllr@{}}
\toprule
Benchmark & Tests & Judge & Items \\
\midrule
AdvBench \citep{zou2023universal} & harm refusal & string match & 520 \\
HarmBench \citep{mazeika2024harmbench} & harm refusal & LLM judge & 296 \\
SORRY-Bench \citep{xie2024sorrybench} & harm refusal & fine-tuned LLM & 438 \\
Do-Not-Answer \citep{wang2023donotanswer} & harm refusal & classifier & 802 \\
AHB \citep{galisai2026adversarialhumanitiesbenchmarkresults} & harm refusal & LLM judge & 676 \\
OR-Bench-Hard \citep{cui2024orbench} & over-refusal & LLM judge & 1{,}318 \\
Health-ORSC \citep{zhang2026healthorsc} & contextual harm & string match & 200 \\
TruthfulQA \citep{lin2022truthfulqa} & truthfulness & BLEURT-20 & 817 \\
\bottomrule
\end{tabular}}
\caption{The eight benchmarks included in our analyses. Item counts are reported after removing unanswered and zero-variance items, leaving 5{,}067 items in total.}
\label{tab:benchmarks}
\end{table}

We collect one response per available model--item pair through OpenRouter's API, using zero temperature and a maximum generation length of 1{,}024 tokens. Analyses involving multiple benchmarks use only models with valid responses on every benchmark required for that analysis.

We evaluate 192 chat models spanning open-weight families from 1B to $\sim$1T total parameters (Llama, Gemma, Qwen, DeepSeek, Mistral, GLM, Kimi, gpt-oss, among others) and proprietary API-only models (GPT, Claude, Gemini, Grok). The 169 models with complete responses on all eight benchmarks form the pool for cross-benchmark analyses; the latent-structure analyses (factor analysis and multidimensional IRT) use the 134 models that additionally have a valid judge score for every retained item on every benchmark (judge failures leave at least one unscored item for each of the remaining 35; Appendix~\ref{app:models} gives the full roster). The complete model inventory and benchmark-coverage matrix are provided in supplementary material.

% The evaluated models come from multiple model families and range from approximately one billion to one trillion parameters. The sample is predominantly composed of open-weight, instruction-tuned models, but also contains base models, reasoning variants, multiple parameter scales, and multiple minor releases from several major model families. For a subset of family--parameter combinations, we evaluate both a base model and its instruction-tuned counterpart.

Each generated response is scored using the judge prescribed by its source benchmark. Let \(x_{mj}\in\{0,1\}\) denote the scored response of model \(m\) to item \(j\), where \(x_{mj}=1\) indicates the response rewarded by the corresponding benchmark.

For harmful-request benchmarks, a value of one generally denotes a refusal. For OR-Bench-Hard, it denotes answering a benign request rather than over-refusing it. For TruthfulQA, it denotes a truthful answer. This orientation means that larger benchmark-specific ability estimates always correspond to higher performance according to the benchmark being fitted.

We remove responses for which the prescribed judge does not return a valid score. We also remove items with no variation across the evaluated models, since these items contain no information about differences within the model cohort. This preprocessing leaves 5{,}067 items.

\subsection{The 2PL IRT Model}
IRT infers a test-taker's latent ability from responses to a fixed set of items. We treat language models as test-takers, benchmark prompts as items, and safe responses as correct answers. The two-parameter logistic (2PL) model is given by
\begin{equation}
P(x_{mj} = 1 \mid \theta_m, a_j, b_j) = \sigma\!\left(a_j(\theta_m - b_j)\right),
\label{eq:2pl}
\end{equation}
where $\theta_m$ is the model's latent ability, $b_j$ is the item's difficulty, and $a_j$ is its discrimination.
Difficulty determines the ability level at which an item changes from usually failed to usually passed (i.e., where $P(x_{mj} = 1) = 0.5$). Discrimination determines how sharply the probability of success changes around that point. Item and ability parameters are inferred jointly by marginal maximum likelihood \citep{baker2004item}; a model's ability is then the posterior mode of $\theta$ under a standard-normal prior, given its responses and the fixed item parameters. Unlike a raw pass rate, this estimate remains on a common scale when different models answer different subsets of items, which enables both short fixed tests and adaptive testing.

We use the 2PL rather than simpler or richer alternatives. The 1PL (Rasch) model constrains all items to discriminate equally, which the data reject: fitted discriminations vary by roughly an order of magnitude within every benchmark, variation that item selection later exploits, and likelihood-ratio tests reject the equal-discrimination constraint on all eight benchmarks. Conversely, the 3PL and 4PL add guessing and slip parameters that improve AIC on only one benchmark and on none, respectively, and BIC never prefers a model richer than the 2PL (Appendix~\ref{app:modelfit}).

Human psychometric datasets tend to have thousands of test-takers, but LLM evaluations contain many more items tested on many fewer models. At this scale, an unregularized maximum-likelihood fit is usually unstable.\footnote{The intuition for this is as follows. For an item that happens to divide the observed models perfectly, the discrimination parameter, $a_j$, won't converge as the closer the logistic gets to a step function, the better the fit. For items that do not discriminate models because they are too hard or too easy, the difficulty parameter, $b_j$, will also tend towards positive or negative infinity. This has downstream consequences for item selection based on these parameters. Small samples of test-takers increase the risk of these problems by reducing the possible variance in response distributions.} We therefore regularize our IRT model by adding a log-normal prior on discrimination and a normal prior on the item intercept. We verify the effectiveness of this using split-half calibration, where the model cohort is divided into two disjoint halves, and the 2PL model is fit separately to each half. Then, we select the 25 items with highest discrimination in the first fit, and measure how much of those items' discrimination is retained in the second. Without regularization, the top-25 items retain only 28\% of their fitted discrimination on AdvBench and 38\% on HarmBench (mean over 10 random splits); with our priors, retention rises to 68\% and 76\%, while the bulk of the parameter distribution is left unchanged.

All results below use the regularized fits\footnote{As IRT tooling defaults can be finicky, cohort sizes like ours will be the norm for LLM work for some time, so such regularization is important to note as best practice.}.

\subsection{Latent Ability Inference}
\label{sec:methods-latents}

We fit a separate 2PL model to each benchmark using the full model cohort. This gives every model one estimated ability per benchmark, producing an $M \times 8$ ability matrix. We then apply minimum-residual factor analysis, which chooses factor loadings that reproduce the observed correlation matrix, with oblimin rotation, which allows the resulting factors to be correlated rather than forcing them to be independent. We select the number of factors using the Root Mean Square Error of Approximation (RMSEA), the Comparative Fit Index (CFI), and Horn's parallel analysis \citep{horn1965parallel, embretson2000item}. Appendix~\ref{app:latent-factor} describes these criteria and the sensitivity of the solution to the extraction method, and Appendix~\ref{app:mirt} reports an item-level multidimensional IRT fit that relaxes the two-stage design.

\subsection{Benchmark Distillation}
\label{sec:methods-distill}

The full benchmark suite before processing contains 5{,}255 items, but many contribute little information about a given model. Easy items are passed by nearly every model, while difficult items are failed by nearly every model. Both consume compute without helping distinguish models at the relevant ability level.

IRT supports two forms of benchmark reduction. First, we can construct a short \emph{static test} containing items that are informative across the model population. Second, we can use \textit{computerized adaptive testing} (CAT) to choose different items for different models according to their estimated ability.

Under the 2PL, the Fisher information of item $j$ at ability $\theta$ is
\[
I_j(\theta)
=
a_j^2 P_j(\theta)\bigl(1-P_j(\theta)\bigr).
\]
An item is most informative near its difficulty, where a model has approximately equal probability of passing or failing, and increases with the square of the item's discrimination. Static tests select items that provide high information across the population's ability range. CAT instead selects the most informative item at the current ability estimate, observes the response, and updates the estimate before choosing the next item \citep{weiss1984application, magis2017computerized}.

We evaluate both methods on held-out models. For each of 20 random splits, we assign 75\% of models to a calibration set and 25\% to a test set. We fit the 2PL and select items using only the calibration models, and evaluate how accurately each reduced test reproduces the full-benchmark results for the held-out models.

Our primary criterion is recovery of the full-benchmark score, measured by root mean squared error (RMSE); we additionally report recovery of model rankings, measured by Spearman correlation, for completeness. To predict full scores, we fit a spline regression on the calibration models using the reduced-test score and estimated ability as inputs.

For comparison, we draw 100 random item sets at each budget. These sets use the same per-benchmark quotas as the informed tests and are passed through the same score-calibration procedure. Differences between informed and random tests therefore reflect item selection rather than test length, benchmark composition, or downstream calibration.

\subsection{Sandbagging Detection and API Auditing}
\label{sec:methods-audit}

IRT models not only estimate ability; they also assign a probability to each response pattern. Person-fit statistics such as $l_z$ compare a model's observed responses with those expected at its estimated ability \citep{drasgow1985appropriateness, meijer2001methodology}. A model that fails easy items while passing difficult ones receives an improbable person-fit score even if its overall pass rate appears normal. In human testing, such patterns can indicate cheating, careless responding, or other departures from the assumed test-taking process.

This provides a basis for auditing individual models using only black-box access. We consider an auditor who has public item parameters and query access to an endpoint, and may additionally have historical ability estimates or reference responses from the claimed model. The auditor is concerned with two failure modes: a model that behaves differently during evaluation than during deployment, and an endpoint that serves different weights than it claims. We evaluate three tests that require progressively more reference information (Table~\ref{tab:ladder}).

\begin{table}[t]
\centering
\resizebox{\columnwidth}{!}{%
\begin{tabular}{@{}lll@{}}
\toprule
Test & Additional requirement & Detects \\
\midrule
Person-fit ($l_z$) & Item parameters & Selective sandbagging \\
Ability shift & Historical $\theta$ & Uniform sandbagging and swaps \\
Response fingerprint & Reference responses & Ability-matched swaps \\
\bottomrule
\end{tabular}%
}
\caption{
Audit methods ordered by the reference information they require. Each method detects failure modes that can remain invisible to the preceding one.
}
\label{tab:ladder}
\end{table}

\paragraph{Person-fit.} We construct controlled prompted-sandbagging interventions on HarmBench's standard-behavior subset (196 items after preprocessing). In each intervention, a system prompt tells the model that it is being evaluated and instructs it to behave with maximum caution. We use three prompt framings (plain, persona, and incentive) and apply each intervention in three scopes: to every item (\emph{uniform}), only to the model's weakest harm category (\emph{targeted}), or only when a trigger phrase appears (\emph{triggered}); we call the targeted and triggered scopes \emph{selective} sandbagging. Detection is fully blind to the intervention: the auditor receives neither labels identifying sandbagged runs nor a clean baseline from the suspect model. A run is flagged when its $l_z$ statistic falls below a threshold set at the 10th percentile of the 171 cohort models with scored responses on this subset, fixing the false-positive rate at 10\%. The 10\% operating point is a policy choice fixing how many genuine models an auditor would needlessly re-examine, not a fitted quantity; Appendix~\ref{app:sandbagging} reports full ROC curves from which any other operating point can be read.

\paragraph{Ability shift.} The second test re-estimates ability from a 40-item test and compares it with the model's historical estimate. A run is flagged when the shift exceeds the smallest ability gap that a 40-item test resolves at that ability level, estimated from pairwise confusability within the cohort. The same procedure quantifies when a substitution of one model for another is detectable from ability alone: for every pair of cohort models, we ask whether a 40-item test distinguishes the pair on at least one benchmark.

\paragraph{Response fingerprint.} The third test compares an endpoint's responses on a pre-selected item set with previously collected responses from the claimed model, and matches each endpoint to the reference model with which its responses agree most. We validate this by re-collecting 16 endpoints on 500 pre-selected items each. Because an endpoint does not reproduce its own responses exactly, we calibrate the expected level of self-disagreement from repeated collections on the same items with the same scoring pipeline (Appendix~\ref{app:sandbagging}).

\section{Results}
\label{sec:results}

\subsection{Latent Ability Inference}
\label{sec:latents}

\begin{figure}[t]
\centering
\includegraphics[width=\columnwidth]{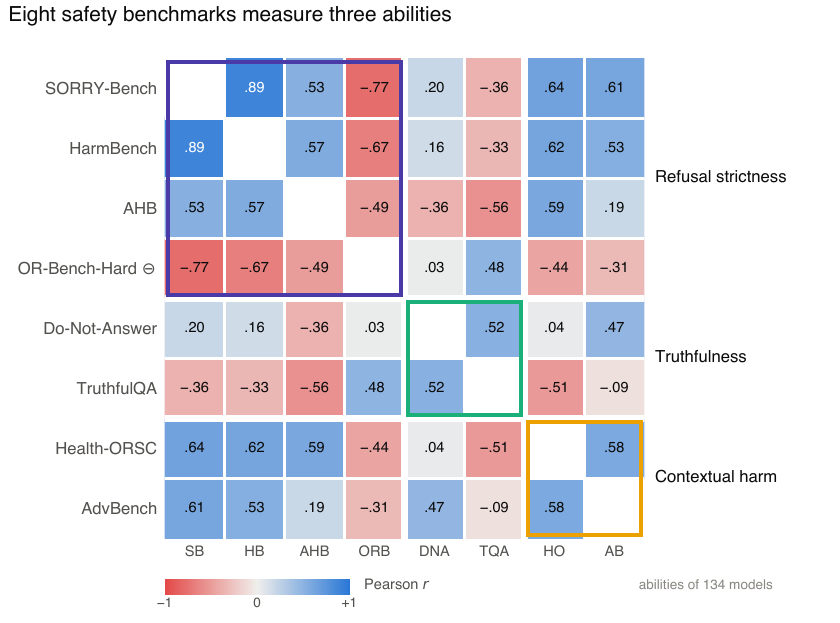}
\caption{
    Pearson correlations between per-benchmark abilities (2PL $\theta$, 134 models), ordered by the three-factor solution. Boxes mark the three ability clusters; OR-Bench-Hard ($\ominus$) correlates negatively with its cluster---it measures refusal strictness reversed. Blue cells outside the boxes reflect correlation between the abilities themselves (refusal $\times$ contextual harm $\phi = .65$); the factors remain distinct because their patterns differ: the harm pair does not share the refusal cluster's over-refusal trade-off (ORB column).
}
\label{fig:structure}
\end{figure}

% Declared here (typeset on the previous page) so the wide grid claims the
% top of the page where the distillation results appear.
\begin{figure*}[tp]
\centering
\includegraphics[width=0.9\textwidth]{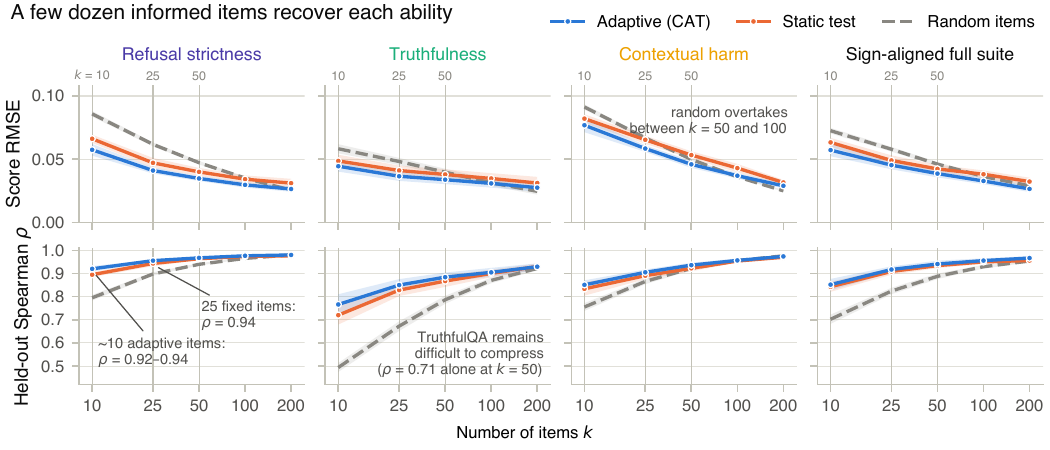}
\caption{
Held-out recovery of the three latent abilities and the sign-aligned full-suite composite from $k$ items, averaged over 20 splits with 95\% confidence intervals. For each target, the upper panel reports score-recovery error (RMSE), our primary criterion, and the lower panel reports rank recovery (Spearman correlation). We compare computerized adaptive testing, a fixed static test, and stratified random subsampling.
}
\label{fig:recovery}
\end{figure*}

\paragraph{Model selection favors three latent factors over one, two, or four.} A one-factor model explains 47\% of the variance in the ability matrix and has poor absolute fit (RMSEA $=0.32$). A two-factor model explains 69\%, with CFI below the conventional threshold for acceptable fit ($0.88$ versus $0.95$). A three-factor model explains 77\% and reaches CFI $=0.97$ (Appendix~\ref{app:latent-factor}, Figure~\ref{fig:factorcount}). Horn's parallel analysis matched to the minimum-residual extraction retains three factors: the third observed eigenvalue exceeds the eigenvalue at the same rank in 1{,}000 random-normal datasets, whereas the fourth does not. The PCA-based variant retains two (Appendix~\ref{app:latent-factor}). We use the three-factor solution throughout.

The three factors group the benchmarks as follows (Figure~\ref{fig:structure}):

\begin{itemize}
\item \emph{Refusal strictness.} HarmBench, SORRY-Bench, and AHB load on one end of this factor, while OR-Bench-Hard loads on the other. HarmBench and SORRY-Bench abilities correlate strongly ($\rho=0.89$), and both correlate negatively with OR-Bench-Hard ($-0.77 \le \rho \le -0.67$).

\item \emph{Truthfulness.} TruthfulQA and Do-Not-Answer define a second factor. This factor is nearly independent of refusal strictness ($\phi=0.01$).

\item \emph{Contextual harm.} AdvBench and Health-ORSC define a third factor through their moderate correlation ($\rho=0.58$). This factor is distinct from refusal strictness, although the two factors are moderately positively correlated ($\phi=0.65$).
\end{itemize}

\subsection{Benchmark Distillation}
\label{sec:efficiency}

\paragraph{Three fixed 25-item tests recover all three latent abilities using under 2\% of the suite.}We construct one 25-item static test for each of the three latent abilities; together they contain 75 items, under 2\% of the full suite. On held-out models, each test estimates its target ability with lower error than a random subset of the same size: RMSE $0.047$ versus $0.062$ for refusal strictness, $0.041$ versus $0.048$ for truthfulness, and $0.065$ versus $0.067$ for contextual harm (Figure~\ref{fig:recovery}). The same tests recover the model orderings at $\rho=0.94$, $0.83$, and $0.89$, against random baselines of $0.90$, $0.67$, and $0.87$. Within the truthfulness test, TruthfulQA is estimated at RMSE $0.052$ against a random baseline of $0.065$; allocating the same number of items across one pooled, all-benchmark test drops its rank recovery from $\rho=0.73$ to $\rho=0.60$.

\paragraph{Roughly ten adaptive items recover individual benchmarks, cutting cost by 97--99\%.}Approximately ten adaptive items estimate HarmBench, SORRY-Bench, and OR-Bench-Hard scores with 18--34\% lower error than a random ten-item subset, corresponding to a 97--99\% reduction relative to administering those benchmarks in full. The same ten items recover their full rankings at $\rho=0.92$--$0.94$. The advantage narrows to 4--21\% at 25 items, and by 50 items random subsampling matches or beats CAT on score recovery for five of the eight benchmarks, although seven still reach $\rho\geq0.90$.

\paragraph{Adaptive testing beats random subsampling on the full-suite composite at every budget.}On the sign-aligned composite of all eight benchmarks, CAT estimates the full score with lower error than random subsampling at every evaluated budget (Figure~\ref{fig:recovery}, right). At 25 items its error is 22\% below a random subset of the same size, and random subsampling does not match it until roughly 55 items; beyond 100 items the two converge to within overlapping confidence intervals. The fixed static test is ahead of random at 10--50 items, and random overtakes it beyond that. Rank recovery shows the same ordering: CAT reaches $\rho=0.92$ at 25 items, a level random sampling reaches at approximately 75--100 items.

% Declared before the subsection so the figure claims the top of the column
% next to the audit results.
\begin{figure}[t]
\centering
\includegraphics[width=\columnwidth]{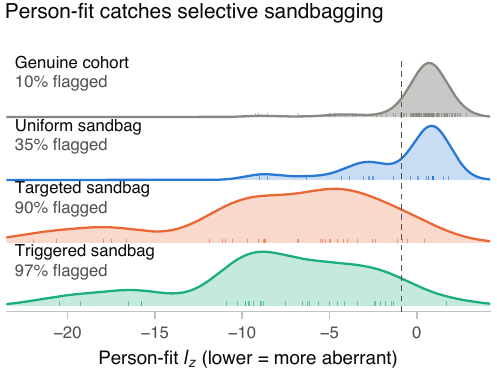}
\caption{
Person-fit $l_z$ distributions on HarmBench for the genuine cohort and the three sandbagging scopes; the dashed line marks the calibrated cutoff (10\% false-positive rate). Selective sandbagging produces aberrant response patterns; uniform sandbagging largely does not (Figure~\ref{fig:auditshift} shows the complementary ability-shift and fingerprint tests).
}
\label{fig:audit}
\end{figure}

\begin{figure}[!t]
\centering
\includegraphics[width=0.825\columnwidth]{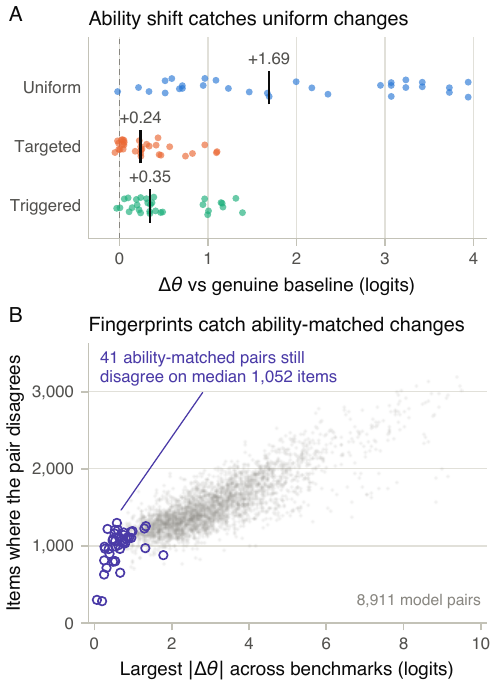}
\caption{
The two audit tests beyond person-fit.
\textbf{(A)} Ability shift $\Delta\theta$ relative to the genuine baseline, by sandbagging scope; uniform sandbagging moves the whole response profile.
\textbf{(B)} Largest per-benchmark ability gap versus total item disagreements for all 8{,}911 cohort pairs; circled points are the 41 ability-matched pairs that a 40-item test cannot separate but that still disagree on a median of 1{,}052 items.
}
\label{fig:auditshift}
\end{figure}

\subsection{Sandbagging Detection and API Auditing}
\label{sec:personfit}

\paragraph{Person-fit catches selective prompted sandbagging but misses most uniform sandbagging.} At the calibrated 10\% false-positive rate, person-fit flags 80--100\% of targeted and triggered sandbagging runs in every prompt framing (90\% and 97\% pooled; Figure~\ref{fig:audit}). Uniform sandbagging is flagged less often: 18--20\% of runs under the plain and persona framings and 70\% under the incentive framing (35\% pooled). Per-framing rates and full ROC curves are given in Appendix~\ref{app:sandbagging}.

\paragraph{Ability shift catches uniform sandbagging; combined, the two tests detect 94--97\% of runs.} Uniform sandbagging changes the estimated ability by a median of $+1.69$ logits and by as much as $+3.94$, whereas targeted and triggered interventions shift ability by only $+0.24$ to $+0.35$ logits (Figure~\ref{fig:auditshift}A). The 40-item ability check against the model's historical $\theta$ flags 84\% of uniform runs, including every plain- and persona-framing uniform run that person-fit misses; combining the two tests detects 94--97\% of sandbagging runs in every scope.

\paragraph{Response fingerprints identify even ability-matched model substitutions.} A 40-item ability test distinguishes 99.5\% of the 8{,}911 model pairs in our 134-model cohort on at least one benchmark. The remaining 41 ability-matched pairs (24 same-vendor snapshots or size variants, 17 cross-vendor pairs) still disagree on a median of 1{,}052 individual items across the full battery, and no cohort pair disagrees on fewer than 284 (Figure~\ref{fig:auditshift}B). Matching the 16 re-collected endpoints to reference responses on 500 pre-selected items identifies 13 of 14 distinct model identities, and correctly declines to distinguish two catalog entries that appear to serve the same weights (Appendix~\ref{app:sandbagging}).

\section{Discussion}

\paragraph{Safety benchmarks measure different abilities.} The eight benchmarks cannot be interpreted as interchangeable measures of a single safety property. For example, HarmBench and SORRY-Bench reward largely the same behavior, while OR-Bench-Hard measures the opposing tendency to answer benign requests rather than refuse them. An unweighted average therefore has two problems. First, it gives extra weight to behaviors measured by several similar benchmarks. Second, it hides the tradeoff between refusing harmful requests and avoiding unnecessary refusals. A model can raise its aggregate safety score by refusing more requests indiscriminately, even when this makes it less useful on benign inputs. Any combined score therefore encodes a substantive choice about how these outcomes should be valued. Evaluators should report distinct abilities when benchmarks separate, justify the weights used in any aggregate, and test whether a new benchmark measures anything not already captured by the suite.

\paragraph{Distilled tests enable frequent evaluation.} Reducing several benchmarks by 97--99\% makes it more practical to evaluate models throughout development rather than only before release. Short tests could be used to compare training interventions, monitor individual checkpoints, and detect changes introduced by fine-tuning, quantization, system prompts, or minor model updates. They could also allow small laboratories and independent auditors to evaluate more models with limited compute. Similar methods may help expand evaluation in low-resource languages and settings, although each new setting would require its own calibration and validation. These efficiencies do not extend the scope of the underlying benchmark. A reduced test shows that the full benchmark can be reproduced with fewer items; it does not show that the benchmark predicts safety in deployment or that the same items will remain informative for future models.

\paragraph{Auditing requires multiple tests.} No single audit detects every kind of behavioral change. Person-fit catches selective prompted sandbagging by identifying response patterns that are inconsistent with a model's estimated ability, but it is less effective against uniform sandbagging, which can resemble ordinary behavior at a different ability level. Historical ability estimates detect these broader shifts, while response fingerprints distinguish models with similar aggregate abilities but different item-level responses. Together, these methods form a practical audit ladder: public item parameters support person-fit, historical estimates reveal drift, and retained responses test whether an endpoint continues to behave like the system originally evaluated. This is especially useful for hosted APIs, where provider routing, quantization, inference backends, and silent serving changes can alter behavior under the same model identifier \citep{khoriaty2026openrouter,pape2026silent,kharinaev2025quantization,gao2025modelequality}. These tests establish behavioral continuity rather than the identity of the underlying weights. Fingerprint comparisons must therefore be calibrated against ordinary variation across repeated measurements of the same endpoint, and our experiments do not show that the methods would detect a model specifically trained to evade them.

\paragraph{Safety evaluation requires psychometric validation.} Safety benchmark scores already inform decisions with consequences, such as decisions to release models or pick one over another. Psychometrics exists because human testing faced the same situation---important decisions resting on noisy, manipulable measurements, such as the case of evaluating psychomotor and cognitive limits for individuals in the military \citep{draheim2025highstakes}. Its response was to make validity, reliability, and test security objects of study in their own right. Our analyses apply the first steps of that program, and none of their conclusions are visible in raw pass rates. The factor names we assign are summaries of loading patterns, not necessarily validated constructs. Using them, or any benchmark score, as grounds for consequential safety claims requires the convergent and predictive validation evidence that psychometrics requires, and safety evaluation has barely scratched the surface. We hope this work encourages both the adoption of these tools and the involvement of researchers trained in measurement.

\section{Limitations}

Our cross-benchmark analyses use up to 169 models, of which 134 enter the latent-structure analyses; this is large for an LLM study but smaller than conventional human-testing datasets that form the basis of most psychometric analysis. This may also influence the interpretation of latent abilities, which we summarize and label using empirical loading patterns but cannot cleanly adjudicate between two or three dimensions. Lastly, our sandbagging detection is evaluated on prompted model organisms rather than fine-tuned ones, which might be easier to catch than scheming models that can vary their responses more adversarially across datasets.

\section{Conclusion}
Psychometrics offers safety evaluation more than compression: a measurement model that says what benchmarks measure, which aggregates are real, how to evaluate at 1--3\% of the cost without losing ranking fidelity, and when an individual model's responses should not be believed. We provide the calibration, the map, the reduced tests, and the audit tools for eight widely used safety benchmarks, and release all artifacts for the community to build on.

% Remove this section for an anonymous AAAI submission.
\section*{Acknowledgments}
We would like to thank Diogo Cruz, Vamshi Bonagiri, Cameron Holmes, and others
for helpful feedback and discussion. We thank the Supervised Program for
Alignment Research for support. We would also like to thank the UK AI Security
Institute and the Department for Science, Innovation, and Technology more
broadly for their support.

\bibliography{aaai2027}

\newpage
\onecolumn
\appendix
% Main sections are unnumbered (secnumdepth 0), which leaves \ref{app:*}
% empty. Restore numbering for the appendix so references render as
% "Appendix A" etc.
\setcounter{secnumdepth}{1}
\section{Models Evaluated}
\label{app:models}
{\small
\setlength{\tabcolsep}{4pt}
\begin{longtable}{@{}l r >{\raggedright\arraybackslash}p{0.72\textwidth}@{}}
\caption{Full model roster (192 models, OpenRouter identifiers with the
developer prefix omitted). $^{\dagger}$~marks the 23 models excluded from
the cross-benchmark analysis cohort (missing from at least one benchmark); the remaining 169 models form the cohort used for
all cross-benchmark analyses. The full response and score data are provided in the supplementary code and data package.}
\label{tab:full-roster} \\
\toprule
Developer & $n$ & Models \\
\midrule
\endfirsthead
\toprule
Developer & $n$ & Models \\
\midrule
\endhead
\bottomrule
\endfoot
Alibaba (Qwen) & 36 & \texttt{qwen-2.5-72b-instruct}, \texttt{qwen-2.5-7b-instruct}, \texttt{qwen-2.5-coder-32b-instruct}, \texttt{qwen-plus}, \texttt{qwen-plus-2025-07-28}, \texttt{qwen2.5-vl-72b-instruct}, \texttt{qwen3-14b}, \texttt{qwen3-235b-a22b}, \texttt{qwen3-235b-a22b-2507}, \texttt{qwen3-30b-a3b}, \texttt{qwen3-30b-a3b-instruct-2507}, \texttt{qwen3-30b-a3b-thinking-2507}, \texttt{qwen3-32b}, \texttt{qwen3-8b}, \texttt{qwen3-coder}, \texttt{qwen3-coder-30b-a3b-instruct}, \texttt{qwen3-coder-flash}, \texttt{qwen3-coder-next}, \texttt{qwen3-coder-plus}, \texttt{qwen3-next-80b-a3b-instruct}, \texttt{qwen3-vl-235b-a22b-instruct}, \texttt{qwen3-vl-30b-a3b-instruct}, \texttt{qwen3-vl-32b-instruct}, \texttt{qwen3-vl-8b-instruct}, \texttt{qwen3.5-122b-a10b}, \texttt{qwen3.5-27b}, \texttt{qwen3.5-35b-a3b}, \texttt{qwen3.5-397b-a17b}, \texttt{qwen3.5-9b}, \texttt{qwen3.5-flash-02-23}, \texttt{qwen3.5-plus-02-15}, \texttt{qwen3.5-plus-20260420}, \texttt{qwen3.6-27b}, \texttt{qwen3.6-35b-a3b}, \texttt{qwen3.6-flash}, \texttt{qwen3.6-plus} \\
\addlinespace[2pt]
Amazon & 3 & \texttt{nova-2-lite-v1}, \texttt{nova-lite-v1}, \texttt{nova-micro-v1} \\
\addlinespace[2pt]
Anthropic & 6 & \texttt{claude-3-haiku}, \texttt{claude-3.5-haiku}$^{\dagger}$, \texttt{claude-haiku-4.5}, \texttt{claude-opus-4}$^{\dagger}$, \texttt{claude-sonnet-4}$^{\dagger}$, \texttt{claude-sonnet-4.5} \\
\addlinespace[2pt]
Arcee AI & 1 & \texttt{trinity-mini} \\
\addlinespace[2pt]
ByteDance & 5 & \texttt{seed-1.6}$^{\dagger}$, \texttt{seed-1.6-flash}$^{\dagger}$, \texttt{seed-2.0-lite}, \texttt{seed-2.0-mini}, \texttt{ui-tars-1.5-7b} \\
\addlinespace[2pt]
Cohere & 3 & \texttt{command-a}$^{\dagger}$, \texttt{command-r-08-2024}, \texttt{command-r7b-12-2024} \\
\addlinespace[2pt]
DeepSeek & 11 & \texttt{deepseek-chat}, \texttt{deepseek-chat-v3}, \texttt{deepseek-chat-v3-0324}, \texttt{deepseek-chat-v3.1}, \texttt{deepseek-r1-distill-llama-70b}, \texttt{deepseek-r1-distill-qwen-32b}$^{\dagger}$, \texttt{deepseek-v3.1-terminus}, \texttt{deepseek-v3.2}, \texttt{deepseek-v3.2-exp}, \texttt{deepseek-v4-flash}, \texttt{deepseek-v4-pro} \\
\addlinespace[2pt]
Essential AI & 1 & \texttt{rnj-1-instruct}$^{\dagger}$ \\
\addlinespace[2pt]
Google & 16 & \texttt{gemini-2.0-flash-001}$^{\dagger}$, \texttt{gemini-2.0-flash-lite-001}$^{\dagger}$, \texttt{gemini-2.5-flash}, \texttt{gemini-2.5-flash-lite}, \texttt{gemini-2.5-flash-lite-preview-09-2025}, \texttt{gemini-2.5-pro}, \texttt{gemini-3-flash-preview}, \texttt{gemini-3.1-flash-lite}, \texttt{gemini-3.1-flash-lite-preview}, \texttt{gemma-2-27b-it}, \texttt{gemma-3-12b-it}, \texttt{gemma-3-27b-it}, \texttt{gemma-3-4b-it}, \texttt{gemma-3n-e4b-it}, \texttt{gemma-4-26b-a4b-it}, \texttt{gemma-4-31b-it} \\
\addlinespace[2pt]
Gryphe & 1 & \texttt{mythomax-l2-13b} \\
\addlinespace[2pt]
IBM & 2 & \texttt{granite-4.0-h-micro}, \texttt{granite-4.1-8b} \\
\addlinespace[2pt]
InclusionAI & 2 & \texttt{ling-2.6-1t}, \texttt{ling-2.6-flash} \\
\addlinespace[2pt]
Inflection & 2 & \texttt{inflection-3-pi}$^{\dagger}$, \texttt{inflection-3-productivity}$^{\dagger}$ \\
\addlinespace[2pt]
Kwaipilot & 1 & \texttt{kat-coder-pro-v2} \\
\addlinespace[2pt]
Liquid AI & 2 & \texttt{lfm-2-24b-a2b}, \texttt{lfm-2-24b-a2b-20260224} \\
\addlinespace[2pt]
Meta & 11 & \texttt{llama-3-70b-instruct}$^{\dagger}$, \texttt{llama-3-8b-instruct}, \texttt{llama-3.1-70b-instruct}, \texttt{llama-3.1-8b-instruct}, \texttt{llama-3.2-11b-vision-instruct}, \texttt{llama-3.2-1b-instruct}, \texttt{llama-3.2-3b-instruct}, \texttt{llama-3.3-70b-instruct}, \texttt{llama-4-maverick}, \texttt{llama-4-scout}, \texttt{llama-guard-4-12b} \\
\addlinespace[2pt]
Microsoft & 3 & \texttt{phi-4}, \texttt{phi-4-mini-instruct}, \texttt{wizardlm-2-8x22b} \\
\addlinespace[2pt]
MiniMax & 6 & \texttt{minimax-01}, \texttt{minimax-m2}, \texttt{minimax-m2-her}, \texttt{minimax-m2.1}, \texttt{minimax-m2.5}, \texttt{minimax-m2.7} \\
\addlinespace[2pt]
Mistral AI & 19 & \texttt{codestral-2508}, \texttt{devstral-2512}, \texttt{devstral-small}$^{\dagger}$, \texttt{ministral-14b-2512}, \texttt{ministral-3b-2512}, \texttt{ministral-8b-2512}, \texttt{mistral-7b-instruct-v0.1}$^{\dagger}$, \texttt{mistral-large}, \texttt{mistral-large-2512}, \texttt{mistral-medium-3}, \texttt{mistral-medium-3.1}, \texttt{mistral-nemo}, \texttt{mistral-saba}, \texttt{mistral-small-24b-instruct-2501}, \texttt{mistral-small-2603}, \texttt{mistral-small-3.1-24b-instruct}, \texttt{mistral-small-3.2-24b-instruct}, \texttt{mixtral-8x22b-instruct}, \texttt{voxtral-small-24b-2507} \\
\addlinespace[2pt]
Moonshot AI & 4 & \texttt{kimi-k2}, \texttt{kimi-k2-0905}, \texttt{kimi-k2.5}, \texttt{kimi-k2.6} \\
\addlinespace[2pt]
Nex AGI & 1 & \texttt{deepseek-v3.1-nex-n1}$^{\dagger}$ \\
\addlinespace[2pt]
Nous Research & 5 & \texttt{hermes-2-pro-llama-3-8b}$^{\dagger}$, \texttt{hermes-3-llama-3.1-405b}, \texttt{hermes-3-llama-3.1-70b}, \texttt{hermes-4-405b}, \texttt{hermes-4-70b} \\
\addlinespace[2pt]
NVIDIA & 4 & \texttt{llama-3.3-nemotron-super-49b-v1.5}, \texttt{nemotron-3-nano-30b-a3b}, \texttt{nemotron-3-super-120b-a12b}, \texttt{nemotron-nano-9b-v2}$^{\dagger}$ \\
\addlinespace[2pt]
OpenAI & 17 & \texttt{gpt-3.5-turbo}, \texttt{gpt-3.5-turbo-0613}$^{\dagger}$, \texttt{gpt-3.5-turbo-16k}, \texttt{gpt-4-turbo}, \texttt{gpt-4.1}$^{\dagger}$, \texttt{gpt-4.1-mini}, \texttt{gpt-4.1-nano}, \texttt{gpt-4o}, \texttt{gpt-4o-mini}, \texttt{gpt-4o-mini-2024-07-18}, \texttt{gpt-5-mini}, \texttt{gpt-5-nano}, \texttt{gpt-5.4-nano}, \texttt{gpt-oss-120b}, \texttt{gpt-oss-20b}, \texttt{o3-mini}, \texttt{o4-mini} \\
\addlinespace[2pt]
Perceptron & 1 & \texttt{perceptron-mk1} \\
\addlinespace[2pt]
Perplexity & 1 & \texttt{sonar} \\
\addlinespace[2pt]
Prime Intellect & 1 & \texttt{intellect-3}$^{\dagger}$ \\
\addlinespace[2pt]
Reka AI & 2 & \texttt{reka-edge}, \texttt{reka-flash-3} \\
\addlinespace[2pt]
Sao10K & 1 & \texttt{l3-lunaris-8b} \\
\addlinespace[2pt]
StepFun & 2 & \texttt{step-3.5-flash}, \texttt{step-3.7-flash} \\
\addlinespace[2pt]
Tencent & 2 & \texttt{hunyuan-a13b-instruct}, \texttt{hy3-preview} \\
\addlinespace[2pt]
TheDrummer & 1 & \texttt{rocinante-12b} \\
\addlinespace[2pt]
Upstage & 1 & \texttt{solar-pro-3} \\
\addlinespace[2pt]
xAI & 3 & \texttt{grok-4.20}, \texttt{grok-4.3}, \texttt{grok-build-0.1} \\
\addlinespace[2pt]
Xiaomi & 3 & \texttt{mimo-v2-flash}$^{\dagger}$, \texttt{mimo-v2.5}, \texttt{mimo-v2.5-pro} \\
\addlinespace[2pt]
Z.ai & 12 & \texttt{glm-4-32b}$^{\dagger}$, \texttt{glm-4.5}, \texttt{glm-4.5-air}, \texttt{glm-4.5v}, \texttt{glm-4.6}, \texttt{glm-4.6v}, \texttt{glm-4.7}, \texttt{glm-4.7-flash}, \texttt{glm-5}, \texttt{glm-5-turbo}, \texttt{glm-5.1}, \texttt{glm-5v-turbo} \\
\end{longtable}}

\twocolumn
\section{Sandbagging and API Auditing Details}
\label{app:sandbagging}

\subsection{Detection Operating Characteristics}

\begin{figure}[t]
\centering
\includegraphics[width=\columnwidth]{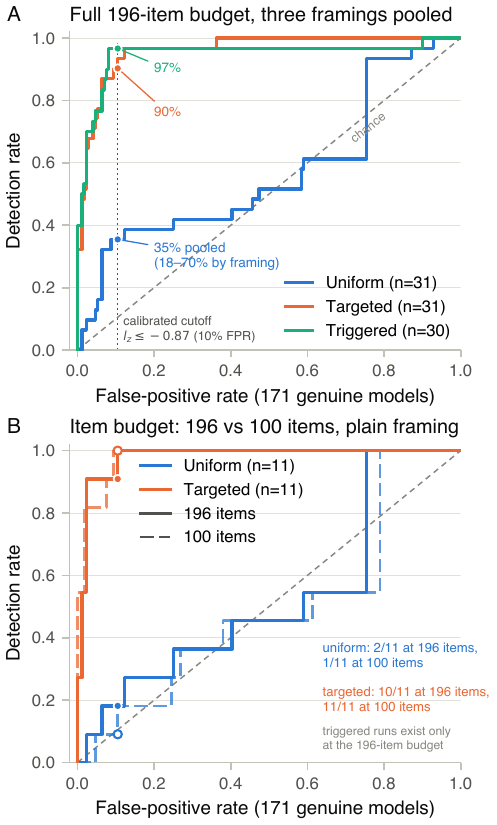}
\caption{
Person-fit detection of sandbagging on HarmBench.
\textbf{(A)} ROC curves per sandbagging scope with the three prompt framings pooled; dots mark the calibrated cutoff ($l_z \le -0.87$, 10\% false-positive rate over 171 genuine models).
\textbf{(B)} The same audit at a 100-item budget (dashed) versus the full 196-item budget (solid), plain framing.
}
\label{fig:auditroc}
\end{figure}

All detection rates in the main-text audit results use a single operating point: a run is flagged when $l_z \le -0.87$, the 10th percentile of the 171 genuine cohort models, i.e.\ a 10\% false-positive rate. Figure~\ref{fig:auditroc}A shows the full ROC curves. By framing (plain, persona, incentive), person-fit flags 91\% (10/11), 100\% (10/10), and 80\% (8/10) of targeted runs; 100\% (10/10), 100\% (10/10), and 90\% (9/10) of triggered runs; and 18\% (2/11), 20\% (2/10), and 70\% (7/10) of uniform runs. The pooled rates in the main text aggregate these unequal run counts.

The detections are not an artifact of the item budget. Repeating the audit with a 100-item subsample of HarmBench's standard-category pool instead of all 196 post-preprocessing items leaves the plain-framing operating points essentially unchanged: targeted runs are flagged at 11/11 (versus 10/11 at the full budget) and uniform runs at 1/11 (versus 2/11; Figure~\ref{fig:auditroc}B).

\subsection{Fingerprint Null Calibration and Drift}

\begin{figure}[t]
\centering
\includegraphics[width=\columnwidth]{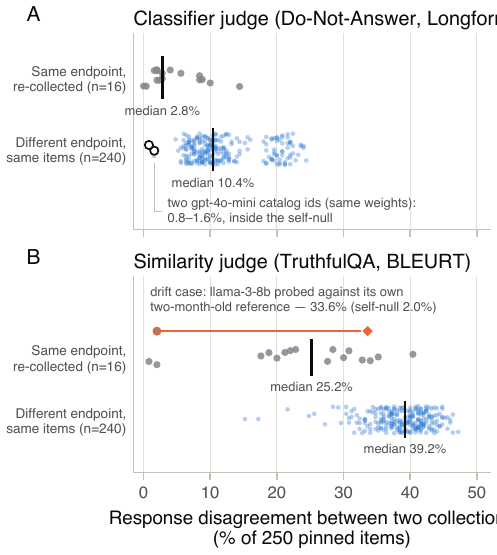}
\caption{
Fingerprint self-disagreement nulls versus cross-model disagreement under a classifier judge (Do-Not-Answer) and a similarity judge (TruthfulQA), with the alias pair and the drift case marked.
}
\label{fig:fpnulls}
\end{figure}

An endpoint does not reproduce its own responses exactly, so fingerprint audits need a null distribution for self-disagreement. Under a stable classifier judge (Do-Not-Answer), endpoints disagree with themselves on a median of 2.8\% of items across repeated collections. Under a threshold-sensitive similarity judge (TruthfulQA), the median rises to 25\%. This variability depends on provider routing, generation stochasticity, item difficulty, and the judging procedure, so the null must be calibrated with repeated collections on the same audit items and the same scoring pipeline (Figure~\ref{fig:fpnulls}).

Against these nulls, cross-model disagreement is well separated from self-disagreement for 13 of the 14 distinct identities in the re-collection experiment. The two catalog entries that the procedure declines to separate agree with one another (8.0--9.4\% disagreement) more closely than either agrees with itself across repeated runs (10.0--10.2\%), consistent with both entries serving the same weights.

The same reference responses support drift audits. One endpoint disagrees with its own reference responses from two months earlier on 33.6\% of TruthfulQA items, against a resampling null of 2.0\%. This change is inconsistent with ordinary endpoint variability under the calibrated null and indicates that the behavior associated with the model identifier changed between collections.

\section{Fitting Multidimensional IRT Models}
\label{app:mirt}

Our main analysis is two-stage. We fit a unidimensional 2PL to each benchmark,
obtain one ability per model per benchmark, and factor-analyze the resulting
$134 \times 8$ ability matrix. A natural alternative is multidimensional IRT
(MIRT): fit $P(x_{mj}=1) = \sigma(a_j^\top \theta_m + d_j)$ with a
vector-valued ability $\theta_m$ directly to the item matrix, so the latent
structure is estimated in one step. This appendix explains why the two-stage
design estimates the same quantity and reports a direct MIRT fit that reaches
the same conclusions.

\subsection{The Two-Stage Design Is a Constrained MIRT Model}

Consider the MIRT model in which each item loads only on its own benchmark's
dimension, with the eight dimensions freely correlated:
$\theta_m \sim \mathcal{N}(0, \Phi)$, where $\Phi$ is the $8 \times 8$
ability correlation matrix. Within a benchmark, responses follow a
unidimensional 2PL; across benchmarks, all dependence runs through $\Phi$.
Under this model, calibrating each benchmark separately loses no information
about item parameters, because the per-benchmark likelihoods are the joint
model's marginals. The second-stage factor analysis then estimates the
structure of $\Phi$ from the ability estimates.

The two-stage estimate of $\Phi$ converges to the truth as the number of
items per benchmark grows, and our benchmarks contribute 200--1{,}318 items
each. The residual measurement error in the ability estimates attenuates the
second-stage correlations toward zero. This biases the analysis
\emph{against} the shared structure we report, so the factor correlations
reported in the main text are conservative. The same separation of
within-benchmark calibration from between-benchmark structure is used by
metabench \citep{kipnis2025metabench}.

One assumption carries the argument: no item loads on more than one ability.
If an item rewarded two abilities at once---say, a harm item that also
rewards truthfulness---the per-benchmark ability would blend the two, and the
second-stage factors could be distorted. An unconstrained MIRT fit tests
exactly this assumption, so we run one.

\subsection{An Unconstrained MIRT Fit Reaches the Same Conclusions}

We fit exploratory MIRT models with unconstrained loadings and
$\theta \sim \mathcal{N}(0, I_D)$ to the $134 \times 5{,}055$ item
matrix (the items retaining response variation within this cohort) at
$D = 1$ to $4$. We compare dimensionalities by 5-fold
cross-validated held-out log-likelihood over models: item parameters are
calibrated on the training folds, and we evaluate the marginal likelihood of
held-out models' complete response patterns without refitting any per-model
parameters. Estimation uses Bock--Aitkin full-information EM
\citep{bock1981marginal} on a pruned Gauss--Hermite grid, echelon constraints
for rotational identification, and a $\mathcal{N}(0, 1.5^2)$ prior on
loadings---the multidimensional analogue of the calibration priors in
Materials \& Methods. The implementation reproduces our unidimensional
2PL exactly at $D=1$ and matches an independent MIRT package on synthetic
two-factor data (loading congruence $1.000$).

\begin{table}[h]
\centering
\begin{tabular}{@{}rrrrr@{}}
\toprule
$D$ & CV loglik & In-sample & AIC & BIC \\
\midrule
1 & $-281{,}521$ & $-266{,}294$ & $552{,}807$ & $582{,}104$ \\
2 & $-266{,}803$ & $-248{,}299$ & $526{,}926$ & $\mathbf{570{,}869}$ \\
3 & $-260{,}009$ & $-237{,}002$ & $514{,}439$ & $573{,}025$ \\
4 & $\mathbf{-257{,}724}$ & $-229{,}606$ & $\mathbf{509{,}751}$ & $582{,}976$ \\
\bottomrule
\end{tabular}
\caption{Item-level MIRT model comparison on the $134 \times 5{,}055$
matrix (bold marks the best value per criterion). Held-out likelihood
improves by $14{,}718$ nats at $D{=}2$ and $6{,}794$ at $D{=}3$, then
flattens ($+2{,}286$ at $D{=}4$); BIC is minimized at $D{=}2$. In-sample AIC
never turns, which is why an out-of-sample criterion is necessary with up to
${\sim}25$k free parameters.}
\label{tab:mirtcv}
\end{table}

\begin{figure}[h]
\centering
\includegraphics[width=\columnwidth]{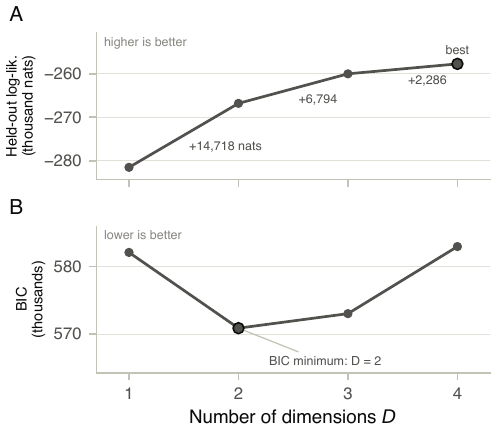}
\caption{Item-level MIRT model comparison: cross-validated held-out
log-likelihood \textbf{(A)} and BIC \textbf{(B)} across latent
dimensionality $D$. Held-out likelihood improves sharply to $D{=}3$ and
only marginally at $D{=}4$; BIC is minimized at $D{=}2$.}
\label{fig:mirtcv}
\end{figure}

The unconstrained fit agrees with the two-stage analysis on the points of
substance (Table~\ref{tab:mirtcv}). Both criteria reject a single dimension
decisively. BIC is minimized at $D=2$, and the held-out gains shrink from
$14{,}718$ nats at $D{=}2$ to $6{,}794$ at $D{=}3$ and $2{,}286$ at $D{=}4$,
consistent with the two-stage conclusion that at least two dimensions are
robust and a third adds modest further structure. Item loadings aggregated
by benchmark recover the benchmark-level structure, including the bipolar
refusal-strictness axis, which appears already at $D=1$ (HarmBench $+0.92$,
SORRY-Bench $+0.90$ versus OR-Bench-Hard $-0.64$). A
permutation parallel analysis \citep{horn1965parallel} on the item
correlation spectrum gives an upper bound of 22 components; permutation
bounds over-count when models are far fewer than items, and the eigenvalue
spectrum itself shows two dominant components with a visible third step.

We report the two-stage analysis in the main text for two reasons. Its
per-benchmark abilities are directly interpretable, and its estimates are
stable at our cohort size, where the unconstrained fit spends up to
${\sim}25$k parameters on 134 response patterns. The item-level fit serves
as the robustness check: the simple-structure constraint is not driving the
conclusions.

\section{Latent Factor Explorations}
\label{app:latent-factor}

\subsection{Factor-Count Selection Criteria}

\begin{figure}[tp]
\centering
\includegraphics[width=\columnwidth]{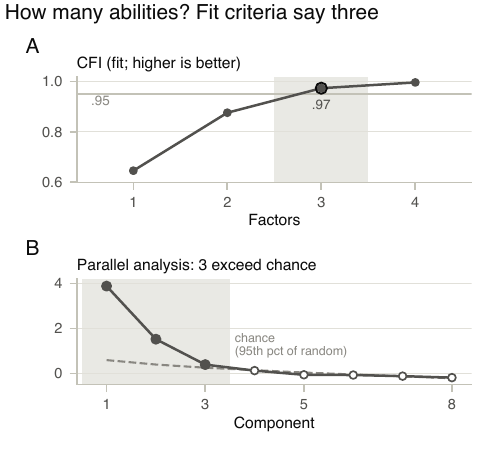}
\caption{
    Model selection for the number of latent abilities.
    \textbf{(A)} CFI first exceeds the conventional $0.95$ threshold at three factors.
    \textbf{(B)} Horn's parallel analysis (minimum-residual): three eigenvalues exceed the 95th percentile of matched random data. Shading marks the selected three-factor solution.}
\label{fig:factorcount}
\end{figure}

We retain factors using three standard criteria. RMSEA measures absolute
misfit of the factor model to the observed correlation matrix, with values
below $0.06$ conventionally taken as close fit. CFI compares the fitted
model against an independence baseline, with values above $0.95$
conventionally taken as acceptable. Horn's parallel analysis retains a
factor when its observed eigenvalue exceeds the 95th percentile of
eigenvalues at the same rank in random data of matched size.

On our eight-indicator ability matrix, the criteria do not agree perfectly.
CFI and minimum-residual parallel analysis both select three factors. The
PCA variant of parallel analysis, which tends toward under-extraction,
retains two. RMSEA improves monotonically with each added factor but reaches
close fit at no solution, a known limitation with few indicators. We
therefore treat multidimensionality, with at least two latent abilities, as
the robust conclusion, and use the three-factor solution to describe the
additional structure supported by CFI and parallel analysis.

\subsection{Composite Scores Require Aligned Directions}

The latent structure also determines which benchmark averages define coherent quantities.

Suppose a composite combines benchmarks that all increase with the same underlying ability. Models that score highly on one member should tend to score highly on the others, and a small subset of items should preserve the resulting model ranking. If some benchmarks point in opposite directions, averaging them cancels part of the shared signal. A related tension appears between AHB and TruthfulQA, whose model abilities correlate at $\rho=-0.56$: high performance on one safety benchmark can therefore predict lower performance on another.

\begin{table}[t]
\centering
\begin{tabular}{@{}lrr@{}}
\toprule
Composite & Signed $\bar{\rho}$ & Recovery @ 25 \\
\midrule
Refusal axis (ORB flipped) & $+0.78$ & $0.96$ \\
\{AdvBench, Health-ORSC\} & $+0.58$ & $0.91$ \\
Harm cluster (5 benchmarks) & $+0.57$ & $0.92$ \\
\{DNA, TruthfulQA\} & $+0.52$ & $0.85$ \\
All 8, raw mean & $+0.10$ & $0.79$ \\
\{ORSC, ORB, TQA\} & $-0.16$ & $0.69$ \\
Refusal trio, raw mean & $-0.18$ & $0.79$ \\
\bottomrule
\end{tabular}
\caption{A composite can be recovered from few items exactly when its members agree in sign. Signed $\bar{\rho}$ is the mean pairwise ability correlation among members (after any flips); recovery is held-out Spearman from 25 adaptively chosen items (20 splits). Rank correlation between the columns: $0.93$.}
\label{tab:composites}
\end{table}

Table~\ref{tab:composites} shows this relationship across seven composites. For each composite, we calculate the mean signed correlation among its constituent benchmark abilities. We then measure how well 25 adaptively selected items recover the full composite. The rank correlation between sign agreement and recovery is $0.93$.

The refusal benchmarks provide a matched comparison. The raw refusal composite combines HarmBench, SORRY-Bench, and OR-Bench-Hard without changing their orientations. Because OR-Bench-Hard measures the opposite end of refusal strictness, its signal partially cancels the other two. The resulting composite has a mean signed correlation of $-0.18$ and a recovery correlation of $\rho=0.79$.

We then reverse OR-Bench-Hard so that all three benchmarks point toward the same end of the refusal-strictness axis. The constituent benchmarks remain unchanged; only the orientation of one score differs. The mean signed correlation rises to $0.78$, and recovery rises to $\rho=0.96$, the highest value among the composites considered.

The same adjustment improves the full eight-benchmark composite. Flipping OR-Bench-Hard before averaging improves every item-selection method we test. Under random subsampling, recovery from 25 items increases from $\rho=0.65$ to $\rho=0.82$. In this setting, correcting the composite produces a larger gain than replacing random sampling with any of the tested item-selection algorithms.

The order of operations therefore matters. Before optimizing which items to administer, one must first determine whether the target score combines compatible dimensions and whether its components are oriented consistently. A more efficient estimate of an incoherent composite remains an incoherent composite.

\section{Why the 2PL}
\label{app:modelfit}

We compare the nested 1PL/2PL/3PL/4PL ladder on each benchmark separately,
fitting each model by marginal maximum likelihood on the same cleaned
matrices used for the main calibration (per-benchmark gap-free models,
post-preprocessing items). The asymptote parameters of the 3PL and 4PL carry
the fitting library's default Beta priors to keep them in the unit interval,
so those two fits are MAP rather than ML and their criteria are approximate;
the 2PL$\to$3PL and 3PL$\to$4PL likelihood-ratio tests also place the
constrained parameter on its boundary, which makes the $\chi^2$ reference
conservative. BIC uses the number of models as $n$, the standard convention
for marginal-likelihood fits.

\begin{table}[t]
\centering
\resizebox{\columnwidth}{!}{%
\begin{tabular}{@{}lrrrrrrr@{}}
\toprule
 & \multicolumn{3}{c}{$\Delta$AIC vs.\ 2PL} & \multicolumn{3}{c}{$\Delta$BIC vs.\ 2PL} & LRT $p$ \\
\cmidrule(lr){2-4} \cmidrule(lr){5-7}
Benchmark & 1PL & 3PL & 4PL & 1PL & 3PL & 4PL & 2PL$\to$3PL \\
\midrule
AdvBench      & $+1005$ & $+1132$ & $+2834$ & $-632$  & $+2772$ & $+6113$  & $1.0$ \\
HarmBench     & $+615$  & $\mathbf{-60}$ & $+895$ & $-304$ & $+863$ & $+2741$ & $8{\times}10^{-29}$ \\
SORRY-Bench   & $+1088$ & $+450$  & $+1090$ & $-280$  & $+1821$ & $+3832$  & $.65$ \\
Do-Not-Answer & $+903$  & $+1066$ & $+3284$ & $-1560$ & $+3532$ & $+8217$  & $1.0$ \\
AHB           & $+578$  & $+603$  & $+2049$ & $-1539$ & $+2723$ & $+6289$  & $.03$ \\
OR-Bench-Hard & $+787$  & $+2880$ & $+5756$ & $-3319$ & $+6990$ & $+13975$ & $1.0$ \\
Health-ORSC   & $+748$  & $+183$  & $+400$  & $+125$  & $+809$  & $+1652$  & $.20$ \\
TruthfulQA    & $+460$  & $+1554$ & $+2997$ & $-2089$ & $+4106$ & $+8102$  & $1.0$ \\
\bottomrule
\end{tabular}%
}
\caption{Model comparison for the nested PL ladder, per benchmark. Negative
values favor the alternative over the 2PL (bold: the single case). The
1PL$\to$2PL likelihood-ratio test rejects the 1PL on every benchmark at
$p < 10^{-109}$ and is omitted from the table.}
\label{tab:modelfit}
\end{table}

Three conclusions follow (Table~\ref{tab:modelfit}). First, the 1PL is
rejected decisively: the likelihood-ratio test rejects the
equal-discrimination constraint on all eight benchmarks, and AIC agrees
everywhere. BIC alone prefers the 1PL on seven benchmarks, which reflects
its penalty of $\log(n)$ per parameter at $n \approx 170$ models applied to
200--1{,}300 additional discrimination parameters rather than a substantive
endorsement: equal discriminations contradict the roughly order-of-magnitude
spread in fitted values, and would eliminate the information weighting that
the item selection in our distillation analyses relies on.

Second, the 3PL improves AIC on exactly one benchmark, HarmBench ($-60$,
LRT $p \approx 10^{-28}$), plausibly because judge leniency on some items
acts like a nonzero floor on the pass probability. The marginal AHB test
($p = .03$) is not corroborated by AIC. Third, the 4PL is never preferred,
and on several benchmarks its fitted likelihood falls below the 3PL's,
which is impossible at a maximum-likelihood optimum; the added asymptotes
make the optimization fail rather than the fit better, the expected outcome
of models with up to ${\sim}5{,}000$ parameters on ${\sim}170$ response
patterns.

The unregularized 2PL and richer fits also reach the iteration cap on most
benchmarks, the divergent-discrimination pathology that motivates the
calibration priors in Materials \& Methods; it affects all rungs above the
1PL alike and does not change any verdict. We therefore use the 2PL
throughout, regularized as described there. A pass-probability floor for
judge-mediated benchmarks may nonetheless be worth revisiting at larger
cohort sizes.

\end{document}